\documentclass[letterpaper, 10 pt, conference]{ieeeconf}  

\IEEEoverridecommandlockouts                              

\usepackage{graphics} 
\usepackage{epsfig} 
\usepackage{mathptmx} 
\usepackage{times} 
\usepackage{amsmath} 
\usepackage{amssymb}  
\usepackage{makecell}
\usepackage{booktabs}
\usepackage{multirow}
\usepackage{subcaption}
\usepackage{acronym}
\acrodef{SNNs}[SNNs]{Spiking Neural Networks}

\title{\LARGE \bf
Embodied Neuromorphic Control Applied on a 7-DOF Robotic Manipulator
}

\author{Ziqi Wang$^{1}$, Jingyue Zhao$^{2}$, Jichao Yang$^{3}$, Yaohua Wang$^{1}$\\ Xun Xiao$^{1}$, Yuan Li$^{1}$, Chao Xiao$^{1}$ and Lei Wang$^{2}$
\thanks{This work was supported in part by the National Natural Science Foundation of China under Grants 62372461,(Corresponding author: Lei Wang)}
\thanks{$^{1}$Ziqi Wang, Yaohua Wang, Xun Xiao, Yuan Li and Chao Xiao are with National University of Defence Technology, Changsha 410071, Hunan, P.R.China. E-mail:
        {\tt\small wangziqi@nudt.edu.cn}}%
\thanks{$^{2}$Jingyue Zhao and Lei Wang are with Defense Innovation Institute, AMS. E-mail:
        {\tt\small leiwang@nudt.edu.cn}}%
\thanks{$^{3}$Jichao Yang is with Qiyuan Lab}
}

\begin{document}

\maketitle
\thispagestyle{empty}
\pagestyle{empty}

\begin{abstract}
The development of artificial intelligence towards real-time interaction with the environment is a key aspect of embodied intelligence and robotics. Inverse dynamics is a fundamental robotics problem, which maps from joint space to torque space of robotic systems. Traditional methods for solving it rely on direct physical modeling of robots which is difficult or even impossible due to nonlinearity and external disturbance. Recently, data-based model-learning algorithms are adopted to address this issue. However, they often require manual parameter tuning and high computational costs. Neuromorphic computing is inherently suitable to process spatiotemporal features in robot motion control at extremely low costs. However, current research is still in its infancy: existing works control only low-degree-of-freedom systems and lack performance quantification and comparison. In this paper, we propose a neuromorphic control framework to control 7-degree-of-freedom robotic manipulators. We use Spiking Neural Network to leverage the spatiotemporal continuity of the motion data to improve control accuracy, and eliminate manual parameters tuning. We validated the algorithm on two robotic platforms, which reduces torque prediction error by at least 60\% and performs a target position tracking task successfully. This work advances embodied neuromorphic control by one step forward from proof of concept to applications in complex real-world tasks.
\end{abstract}

\section{INTRODUCTION}
The transition of artificial intelligence from offline learning based on fixed datasets to real-time interaction with the environment during operation represents a key aspect of embodied intelligence \cite{hughes2022embodied}. The implementation of embodied intelligence in the field of robotics holds significant importance, as it endows robots with heightened adaptability to their surroundings and enhanced autonomous decision-making capabilities \cite{bartolozzi2022embodied, roy2021machine}. The inverse dynamics problem is fundamental for enabling robots to perform complex tasks. It requires the calculation of the torques needed at each joint to achieve the robot's desired state.

Inverse dynamics physical modeling is the process of deriving the forces and torques applied to a system using physical principles and equations based on known motion states. It is difficult or even impossible to directly build physical models of inverse dynamics due to external disturbances and difficulties in accurately obtaining parameters \cite{nguyen2010using}. In recent years, data-based model learning algorithms have become popular for modeling complex robotic systems \cite{polydoros2015real}. These algorithms offer numerous advantages in control, such as potentially higher tracking accuracy, reduced feedback gains, and more compliant control \cite{reinhart2017hybrid}. For instance, Vijayakumar et al. \cite{vijayakumar2005lwpr} introduced Locally Weighted Projection Regression (LWPR)  that uses local linear models to approximate nonlinear functions in high-dimensional space. However, the algorithm requires manual tuning of many data-dependent parameters. Furthermore, there are also many methods that use kernel functions. Non-parametric regression methods like Gaussian Process Regression (GPR) and Support Vector Regression (SVR) are easy to use, but their computational complexity grows exponentially in $O(n^3)$ \cite{nguyen2008learning}. PC-ESN \cite{polydoros2015real} uses Artificial Neural Network (ANN) to solve the 7-degree-of-freedom (7-DOF) inverse dynamics problem, but ANN treats each sample independently and cannot utilize the spatiotemporal continuity between samples. Therefore, there is a lack of a model learning algorithm that both reduces manual parameter tuning and computational complexity while fully utilizing the spatiotemporal continuity between motion data.

\begin{figure}[t]
  \centering
  \includegraphics[width=0.5\textwidth]{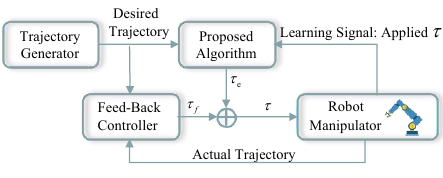}
  \caption{Flowchart of Trajectory Tracking Application. The algorithm predicts the torque based on the desired trajectory (joint positions, velocities, and accelerations), and the feedback controller corrects the predicted torque. Sensors read the actual torque applied to the robotic arm and feed it to the algorithm as a learning signal.}
  \label{pic_intro}
\end{figure}

The development of neuromorphic computing offers a new approach to solving inverse dynamics problems. In particular, \ac{SNNs} mimicking the characteristics of the human brain, use discrete spike trains to transmit information between neurons. Due to the continuity of membrane potential accumulation along the temporal dimension, each neuron can be regarded as a dynamic system. The network made of such neurons has a powerful ability to memorize and detect spatiotemporal information \cite{fang2020exploiting}. Additionally, Benefiting from the event-driven property, SNNs also are energy efficient on neuromorphic hardware \cite{liu2021event}. Therefore, SNNs have broad application prospects in the field of robotic control \cite{zhao2020closed}. A feed-forward SNN is proposed to tackle the inverse dynamics problems of a 3-DOF robotic arm \cite{kapatsyn2024real}. Gilra et al. achieved trajectory tracking of a two-link system by training a heterogeneous SNN \cite{gilra2018non}. However, these works control relatively low (i.e., only 2 or 3) degrees of freedom. Moreover, they all belong to feedforward networks, which struggle to handle highly nonlinear tasks. Compared to the liquid state machine (LSM) with recurrent connections used in this paper, they are less capable of retaining the historical state of previous inputs, which hinders them from complex robotic systems and practical applications.

In this paper, we delve into solving the inverse dynamics problem for a 7-DOF robotic arm using SNN. 
The general process is illustrated in Fig. \ref{pic_intro}. We leverage SNNs' inherent spatiotemporal information processing capability to effectively capture the spatiotemporal continuity feature in the robot motion, which improves the control precision of the movements. Additionally, We conducted control experiments on different robot platforms, the Baxter robot \cite{fitzgerald2013developing} and the iCub humanoid robot \cite{metta2008icub}, demonstrating the effectiveness of the neuromorphic control algorithm. The main contributions of this work are as follows.

\begin{itemize}

\item We present a neuromorphic control framework that solves the inverse dynamics problem and use it to perform torque control on two 7-DOF robotic arms. The framework enhances system stability and accuracy by exploiting the spatiotemporal continuity of inverse dynamics data on the physical dynamics.
\item We propose an SNN-based model for continuous torque prediction given time-varying robot states. This model uses delta encoding and an LSM to extract and retain the spatiotemporal features of robot trajectory without requiring extensive manual parameter tuning. 
\item We make publicly available an embodied control dataset obtained from the iCub robot specifically for the inverse dynamics learning problem, which eases the lack of datasets based on the robot's proprioception to boost the emerging research of embodied intelligence.
\end{itemize}

The experimental results show that this method reduces the average error of two 7-DOF robotic arms by over 60\% compared to other algorithms, and by up to 36\% compared to other encoding schemes. When applied to the iCub trajectory tracking task, the method enables the iCub robotic arm to effectively track the desired trajectory.

\section{PRELIMINARIES AND RELATED WORK}
\subsection{Spiking Neuron and Synapse}
\begin{figure}[t]
  \centering
  \includegraphics[width=0.5\textwidth]{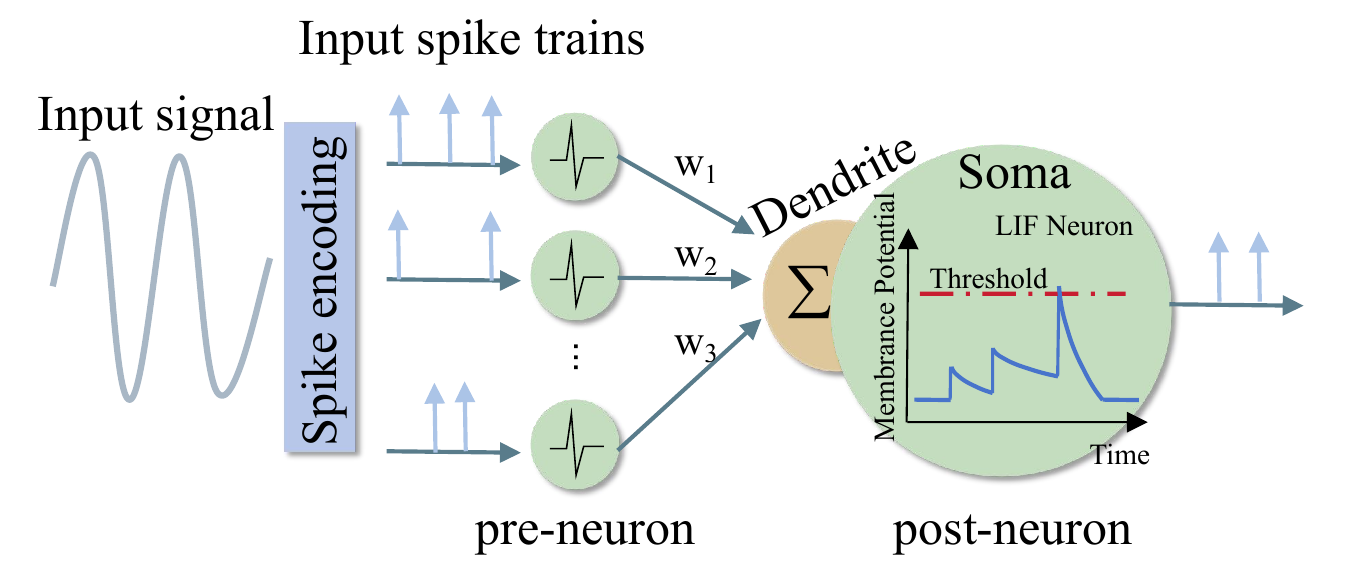}
  \caption{Single Neuron Simulation. The post-neuron receives and integrates the current transmitted from the pre-neuron, with its membrane potential gradually accumulating over simulated time steps until it fires a spike and returns to the resting potential.}
  \label{pic-neuron}
\end{figure}

SNNs are a type of ANNs inspired by biological nervous systems. In contrast to traditional ANNs, SNNs transmit information through spike trains. Each spike represents a binary signal, where 0 indicates no spike is fired, and 1 indicates a spike is fired. Neurons in SNNs are divided into two types: excitatory neurons and inhibitory neurons. When excitatory neurons fire spikes, they increase the membrane potential of the postsynaptic neuron, while inhibitory neurons have the opposite effect. Similar to biological neurons, when the neuron's voltage exceeds a threshold, the neuron fires a spike, and its membrane voltage is reset to the resting potential. Fig. \ref{pic-neuron} illustrates the activity of a single neuron. Before a neuron fires a spike, the membrane potential accumulates over time, which is influenced not only by the current input but also by the retention of past information. This characteristic gives SNNs a natural ability to process spatiotemporal data.

To date, many neuron models have been proposed to simulate the behavior of real neurons. The Hodgkin-Huxley (HH) model is the earliest biological model. Although the HH model is highly accurate biologically, it has high computational costs. The leaky integrate-and-fire (LIF) model was proposed as a simplified model. While retaining the basic dynamical behavior of neurons, it significantly reduces computational complexity by linearizing the changes in membrane potential and introducing a leakage term. In particular, the equations describing the model are given by:
\begin{equation}
    \frac{dV}{dt}~=~\frac{-(V-V_{rest})}{\tau_1}~+~\frac{I}{\tau_2}
    \label{c}
\end{equation}
where $V$ is the membrane potential and $V_{rest}$ is the resting potential. Both $\tau_{1}$ and $\tau_{2}$ are the membrane's temporal charging constant. $I$ is the input current received by the neuron. When a neuron receives a spike input, the input current $I$ increases based on the synaptic strength and subsequently decays over time. The update rule for $I$ is as follows:
\begin{equation}
    \frac{dI}{dt}~=~-\frac I{\tau_3}~+~\sum W_is_i
    \label{d}
\end{equation}
where $\tau_{3}$ represents the time constant for the decay of the input current. $\sum W_is_i$ is the weighted sum of inputs from all presynaptic neurons. 

\subsection{Liquid State Machine}
LSM is a type of Recurrent Neural Network (RNN) based on spiking neurons \cite{maass2002real}. As shown in Fig. \ref{pic_NN}b, LSM primarily consists of an input layer, a liquid layer, and a readout layer. The connections between the input layer and the liquid layer are random, and the input layer is only connected to the excitatory neurons in the liquid layer. The liquid layer is the main component of the LSM, where random connections exist between excitatory and inhibitory neurons. These include self-connections as well as connections to other neurons. The current into an excitatory neuron k in the liquid layer is given by
\begin{equation}
J_k(t)=\sum_n w_{kn}s_n(t) + \sum_m w_{km}s_m(t-1)
\label{h}
\end{equation}
$\sum_n w_{kn}s_n(t)$ is the input of the input layer at the current time step. $\sum_m w_{km}s_m(t-1)$ is the input to the liquid layer from itself at the previous timestep.

Due to the recurrent connections and nonlinear activation among spiking neurons, LSM has the ability to extract features, dynamically remember and utilize past information, and better handle time series data. In this work, features are extracted by recording the number of spikes fired by the neurons. 

\subsection{Related Work}
In recent years, many improved machine-learning methods have been proposed. Nguyen-Tuong et al. \cite{nguyen2009model} proposed a local Gaussian process (LGP) regression method, combining the advantages of both GPR and LWPR, offering high accuracy and real-time performance. Subsequently, they further improved the SVR by applying sparsification, which leverages the linear independence of the training data \cite{nguyen2009sparse}. A neural network and GPR are combined in \cite{hartmann2013real}, an RNN to represent the state of the robot arm, and novel online Gaussian processes for regression. Rueckert et al. \cite{rueckert2017learning} used long-short-term-memory (LSTM) network to learn the inverse dynamics model, and this approach has lower computational complexity.

\section{METHOD}
\subsection{Problem Definition}
For a N-DOF robotic arm, the inverse dynamics can be modeled as:
\begin{equation}
   \tau~=~M(q)\ddot{q}~+~C(q,\dot{q})~+~G(q)
   \label{a}
\end{equation}
where $q,\dot{q},\ddot{q}\in\mathbb{R}^n$ are the joint positions, velocities and accelerations respectively \cite{patel2021deep}. $M(q)\in\mathbb{R}^{n\times n}$ is the inertia matrix. $C(q,\dot{q})\in\mathbb{R}^{n\times n}$ is the Coriolis and Centripetal matrix. Finally, $G(q)$ is the forces or torques generated by gravity acting on the system and $\tau$ refers to the joint torque vector. In actual systems, the precise parameters for each term are often difficult to accurately obtain, and the system may be affected by external disturbances such as friction, collisions, and other environmental factors \cite{nguyen2011model}. This makes defining physics-based models a challenging task.

The objective of this work is to learn the inverse dynamics model described in Eq. \ref{a}. We approximate this problem as a regression task. Given the desired trajectory $\boldsymbol{x}$ and the corresponding torque $\boldsymbol{\tau}$, the goal is to learn the function $f(x)$ as described in Eq. \ref{b}.
\begin{equation}
\boldsymbol{\tau}~=~f(\boldsymbol{x})~+~\boldsymbol{\epsilon}~:~\mathbb{R}^{3N\times1} \mapsto \mathbb{R}^{N\times1}
\label{b}
\end{equation}
where $\epsilon$ denotes the Gaussian noise of the system.


Due to the temporal continuity of the robot joints' motion, the state variables such as the joint positions and velocities change continuously between adjacent moments. Specifically, the state at each moment depends not only on the current control input but also on the previous state. Additionally, the movement of the robotic arm is smooth rather than abrupt. Therefore, the torque applied to the joints also exhibits temporal correlation \cite{rueckert2017learning}.

In data-based inverse dynamics learning, temporal continuity is particularly important. Since the system's historical state can provide valuable predictive information for the current state, the network can leverage this characteristic to enhance the model's robustness against nonlinearity and external disturbances, thereby improving the accuracy and smoothness of inverse dynamics predictions, especially when performing precise control in complex environments.

\begin{figure}[t]
  \centering
  \includegraphics[width=0.5\textwidth]{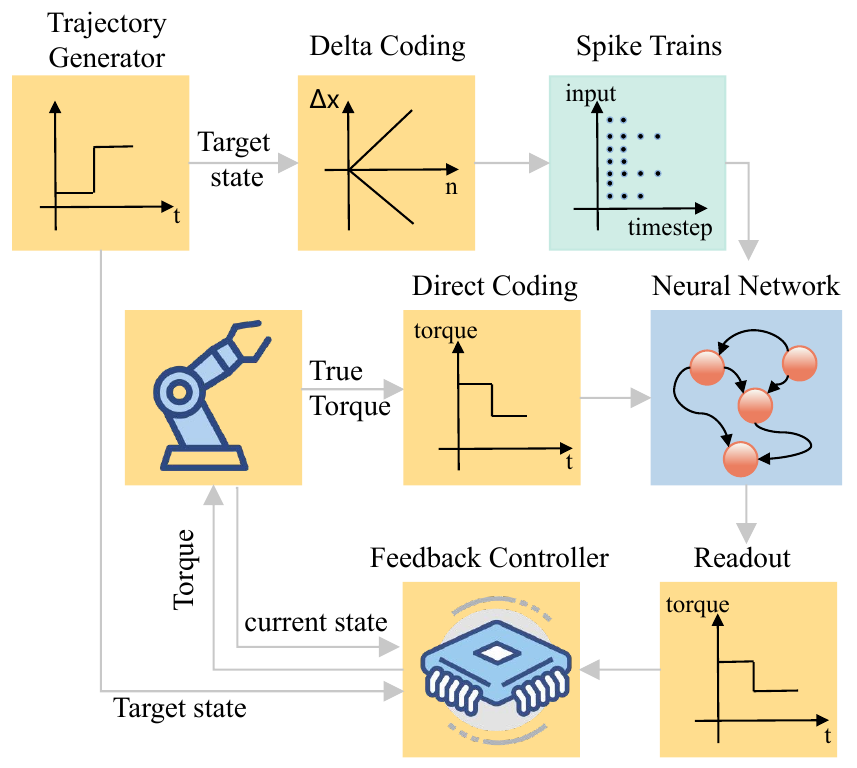}
  \caption{Overall Control Framework. The control framework adopts closed-loop control, where sensors on the robotic arm output the current state and the actual torque applied to the arm, ensuring that the robotic arm accurately reaches the target state and tracks the desired trajectory.}
  \label{pic_overrall}
\end{figure}

\begin{figure*}[htbp] 
    \centering
    \begin{minipage}[b]{0.35\textwidth}
        \centering
        \includegraphics[width=\textwidth]{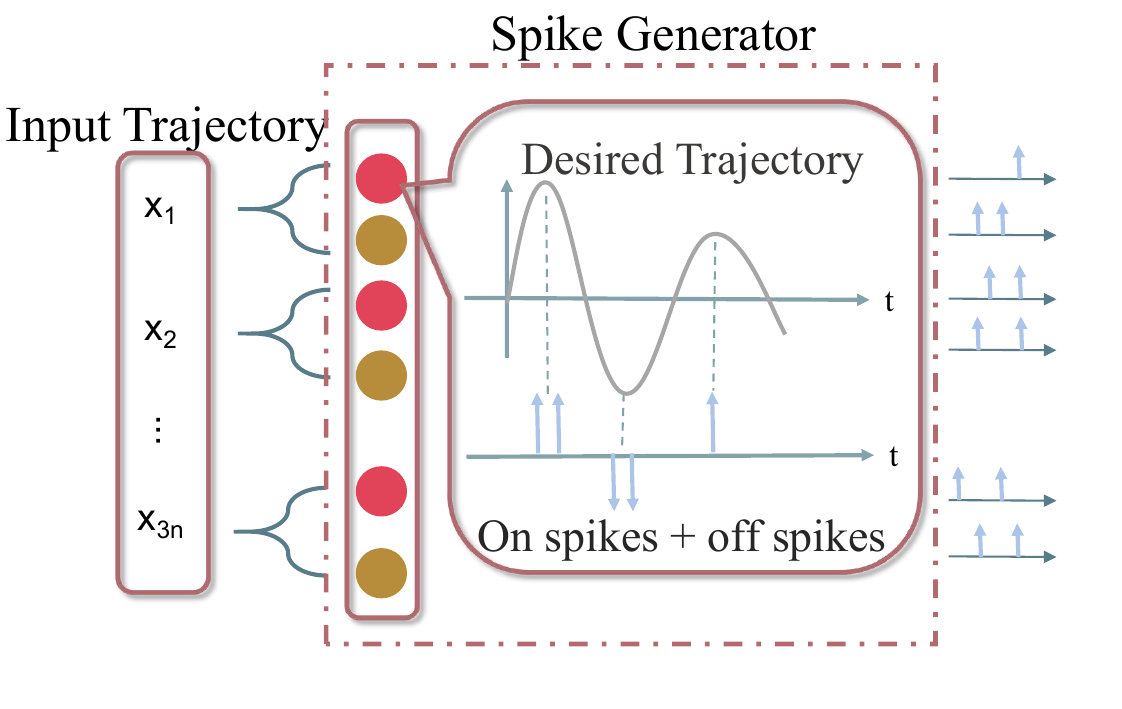}
         \subcaption{Spike Encoding}
    \end{minipage}
    \hfill
    \begin{minipage}[b]{0.62\textwidth}
        \centering
        \includegraphics[width=\textwidth]{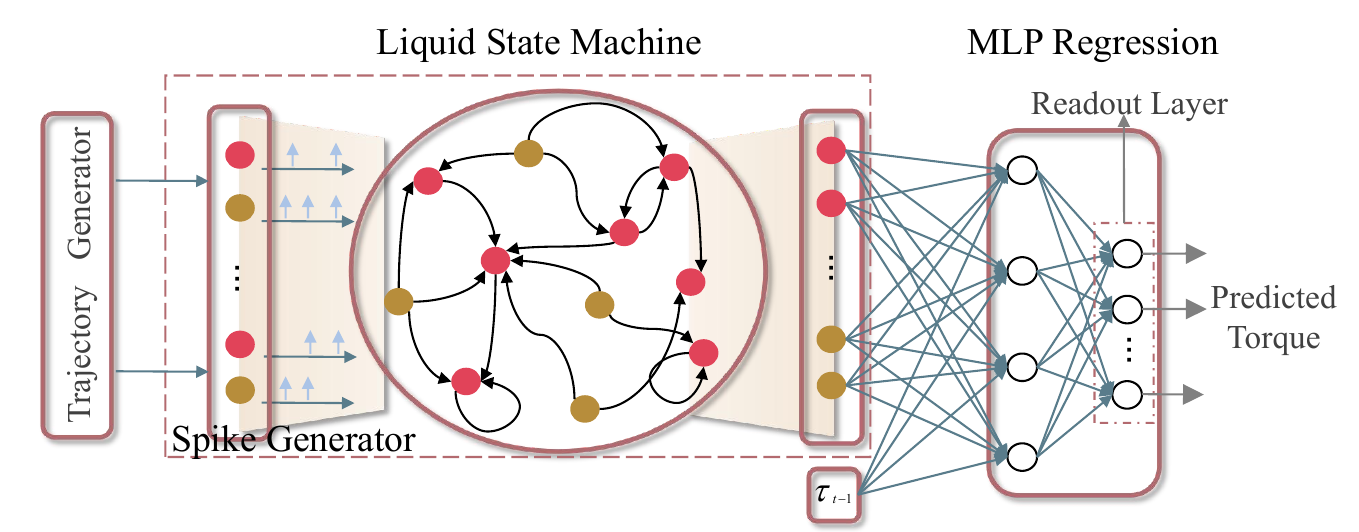}
        \subcaption{Neural Network}
    \end{minipage}
    \caption{(a) Convert the input trajectory values into spikes using delta encoding. If the change is positive, excitatory neurons (red circles) fire spikes; otherwise, inhibitory neurons (yellow circles) fire spikes. (b) The LSM extracts and retains the feature values of the target robotic arm's state through its internal recurrent connections, and sends the feature values to the MLP. The MLP combines the torque value applied to the robotic arm in the previous time step to predict the torque.}
    \label{pic_NN}
\end{figure*}

\subsection{Overall Control Framework}
The overall framework of the control system is shown in Fig. \ref{pic_overrall}. The trajectory generator produces the desired state for the robotic arm at the next time step (joint positions, velocities, and accelerations), which is converted into spike inputs via delta encoding and fed into the neural network. The network's readout layer outputs the predicted torque. A feedback controller then adjusts the torque based on the current state and the desired state to ensure precise trajectory tracking of the robotic arm. Finally, the actual torque applied to the robotic arm is directly encoded and provided to the network to assist in predicting the torque for the subsequent moment. Improving the network's prediction accuracy can effectively reduce the feedback gain of the feedback controller, leading to a more stable operation of the entire system.

\subsection{Network Architecture}
In this section, we will introduce the composition of the network, followed by a detailed description of each module.

Fig. \ref{pic_NN}b shows the neural network that predicts the torque values for each degree of freedom to achieve the desired state. The network consists of two parts:
\begin{itemize}
    \renewcommand{\labelitemi}{} 
    \item LSM: Map the target state encoded by delta spike encoding into a high-dimensional space to extract and retain the features of the robotic arm's target state.
    \item Multilayer Perceptron (MLP) Regression: Performs regression on the extracted features and outputs the predicted torque $\vec{u}(t)$ that generated the arm state trajectory $\vec{x}(t)$.
\end{itemize}
%

The connections between the input neurons and the liquid neurons and the internal connections of the liquid layer are constant, which can reduce training complexity and accelerate the convergence of the model. The connections between the LSM and the MLP are feed-forward and they are updated according to the backpropagation rule. 

\subsubsection{Spike Encoding} 
Here, we adopt the Delta compression coding method \cite{whitley1991delta}, as shown in Fig. \ref{pic_NN}a. The data at the first moment is unchanged, and the other data compares the input at the current moment with the input at the previous moment (Eq. \ref{e}). The 21-dimensional state of the arm (7 joint angles, 7 joint velocities, and 7 joint accelerations), is fed as input to the network. Each dimensional state input corresponds to two neurons, with each neuron encoding different information. If the change is positive, the excitatory neuron will fire. Otherwise the inhibitory neuron fires. This encoding scheme transmits spikes only when there is a change in the signal, thereby saving computational and storage resources. Additionally, compared to traditional encoding methods (such as rate encoding \cite{theunissen1995temporal} and direct encoding \cite{gruau1996comparison}), it can effectively capture the dynamic changes in the robotic arm motion data, making it more suitable for addressing inverse dynamics problems characterized by spatiotemporal continuity.
\begin{equation}
    \Delta x(t)~=~x(t)~-~x(t-1)
    \label{e}
\end{equation}

The network is equipped with hyperparameters $\theta$ to adjust the number of spikes fired by neurons:
\begin{equation}
    n~=~\Delta x~/~\theta
    \label{f}
\end{equation}
\begin{equation}
   n~=~\begin{cases}timestep&n\geq timestep\\n&n<timestep\end{cases} 
   \label{g}
\end{equation}
where $n$ represents the number of spikes that the neuron fires.

\subsubsection{LSM Structure and Parameter Search}
Approximately 100 liquid neurons are used in our LSM model. We randomly select some of them as inhibitory neurons and the rest as excitatory neurons. LSM neurons form a cube structure. We define the distance between neurons i and j according to (\ref{j}), and then use the distance to calculate the connection probability between neurons i and j according to (\ref{k}). It is easy to conclude that the smaller the distance between neurons, the higher the probability of connection between neurons. This aspect also improves the biological plausibility of the network.

\begin{equation}D_{i,j}~=~\sqrt{(x_i-x_j)^2+(y_i-y_j)^2+(z_i-z_j)^2}
\label{j}
\end{equation}
\begin{equation}P_{i,j}~=~C~\times~e^{-(D_{i,j}/\lambda)}
\label{k}
\end{equation}

\begin{table}[ht]
        \newcommand{\tabincell}[2]{\begin{tabular}{@{}#1@{}}#2\end{tabular}}  
	\caption{Parameter Search Range}
        \centering
        \setlength{\tabcolsep}{15pt} 
        \renewcommand{\arraystretch}{1.3} 
	{\begin{tabular}{c|c|c}  	
		\hline	
		Name& Search range& Type\\ \hline
		$\text{n}$ & \makecell[c]{100 to 512 }  & int\\

$P_{\text{input}}$ & 0.25 to 0.85 & float \\

$C_{\text{ee}}$ & 0.4 to 1.7 & float\\

$C_{\text{ei}}$ & 0.2 to 1.5 & float\\

$C_{\text{ie}}$ & 0.2 to 1.5 & float\\

$C_{\text{ii}}$ & 0.1 to 0.7 & float\\
\hline
	\end{tabular}} 
	\label{Parameter}
 
\end{table}

Since the structure of LSM remains unchanged after initialization and different structures have different capabilities for feature extraction, we used the Particle Swarm Optimization(PSO) algorithm to search the connection probability of LSM that is more suitable for trajectory tracking tasks during initialization. The parameters to be searched in the network include $n$ (number of neurons in the liquid layer), $P_{input}$ (Probability of connection between input neurons and the excitatory neurons in the liquid layer) and $C$ (Probability of connections in the liquid layer). $C$ has four values: $C_{ee}$ between excitatory neurons and excitatory neurons, $C_{ei}$ between excitatory neurons and inhibitory neurons, $C_{ie}$ and $C_{ii}$, and so on. The scope of the parameter search is shown in table \ref{Parameter}.

\section{EXPERIMENTS}
\subsection{iCub Dataset}
\begin{figure}[h]
  \centering
  \includegraphics[width=0.45\textwidth]{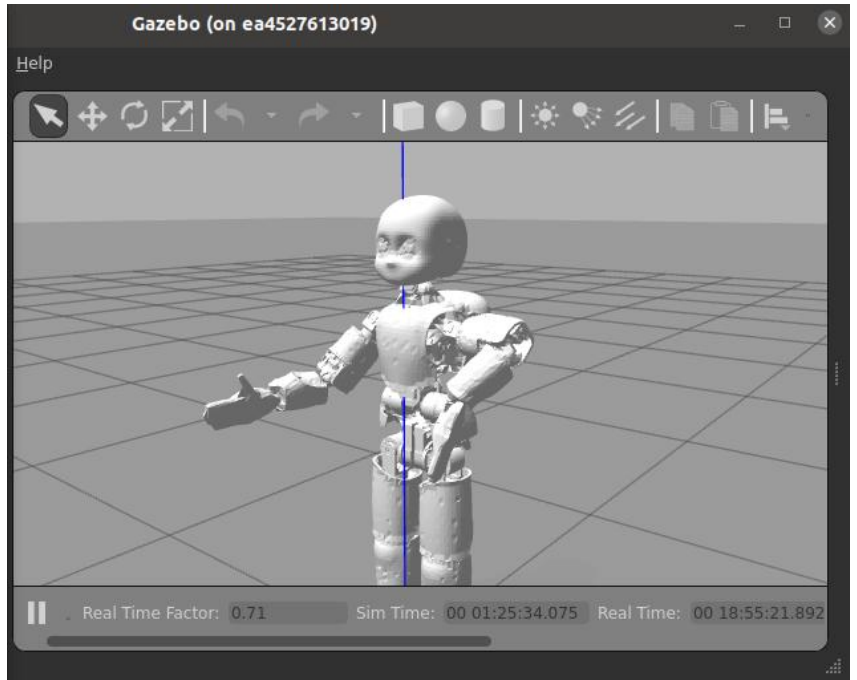}
  \caption{iCub Robot Simulator}
  \label{pic_icubsimulator}
\end{figure}

To evaluate the proposed controller architecture, we utilized the open-source iCub humanoid platform simulator Gazebo as a test platform, controllable using the Yet Another Robot Platform (YARP) middleware \cite{metta2006yarp}, as shown in Fig. \ref{pic_icubsimulator}. This configuration facilitates the simulation and evaluation of the system on a virtual platform, while also enabling deployment to the physical iCub robot using the same interface. In this work, the system uses a torque control mode to manage the 7-DOF left arm of the iCub robot.

Our data collection process for the iCub 7-DOF robotic arm is similar to that of the Baxter robot, with data generated by point-reaching movements. We sampled the joint state positions, velocities, and accelerations, and applied torque at each joint at a frequency of 100 Hz using the joint\_encoders and force-torque sensors on the robot's left arm. We collected a total of 7,958 data points, with the first 5,271 used as training data and the remaining data used for testing. We have made publicly available this dataset at: https://bitbucket.org/icubdataset/inverse-dynamic/downloads/

\subsection{Control Accuracy}
\subsubsection{Performance on Baxter Robot}
\begin{figure}[h]
  \centering
  \includegraphics[width=0.48\textwidth]{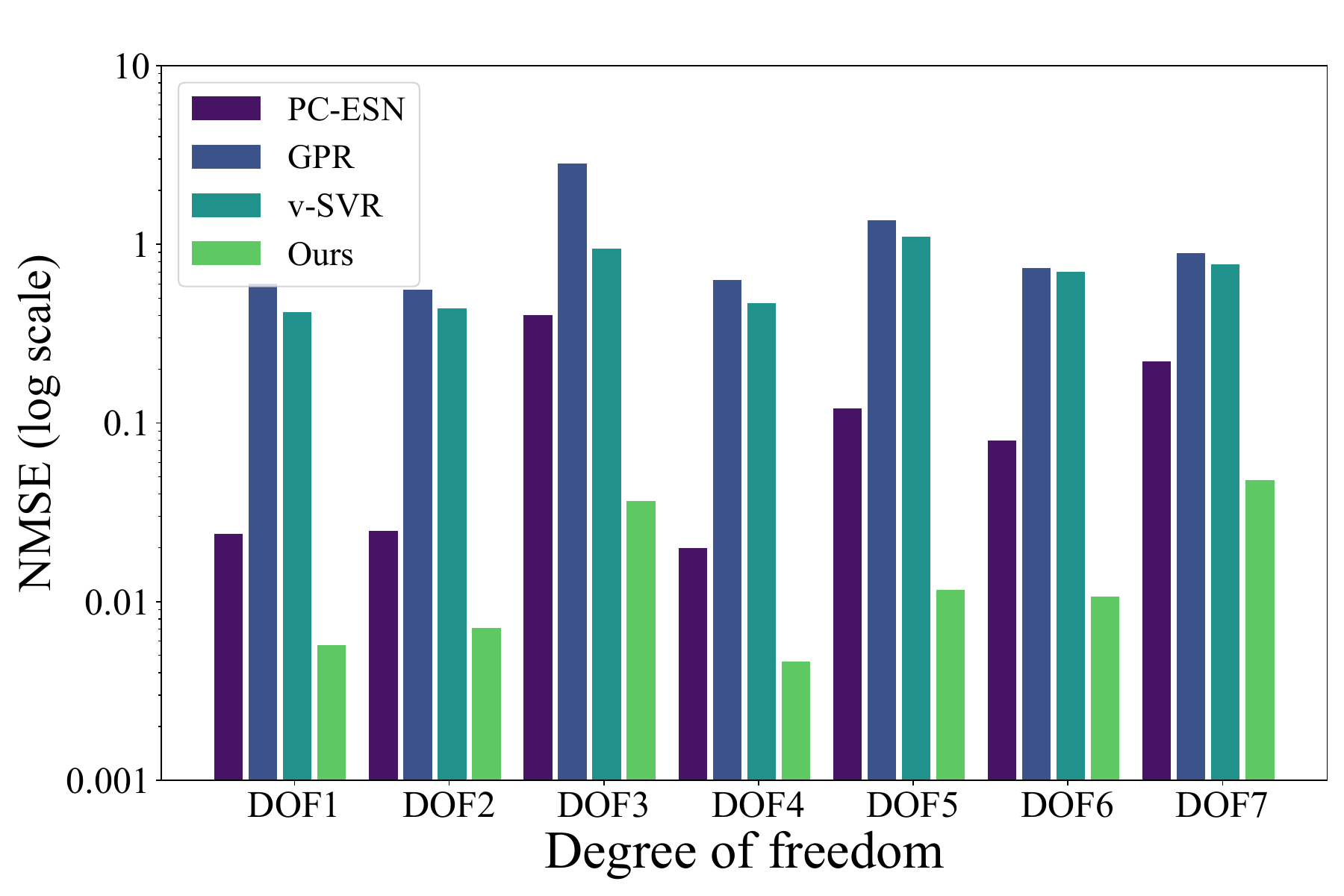}
  \caption{Accuracy on the Baxter Robot}
  \label{pic_baxternmse}
\end{figure}

In Fig. \ref{pic_baxternmse} We compared the proposed algorithm to three other algorithms on the Baxter robot. The results demonstrate that our method can provide more accurate predicted torque. The figure shows the experimental results of PC-ESN with the reservoir size set to 100. The minimum average NMSE for PC-ESN is 0.05, which is still higher than that of our method.

\subsubsection{Performance on iCub Robot}
\begin{figure}[h]
  \centering
  \includegraphics[width=0.48\textwidth]{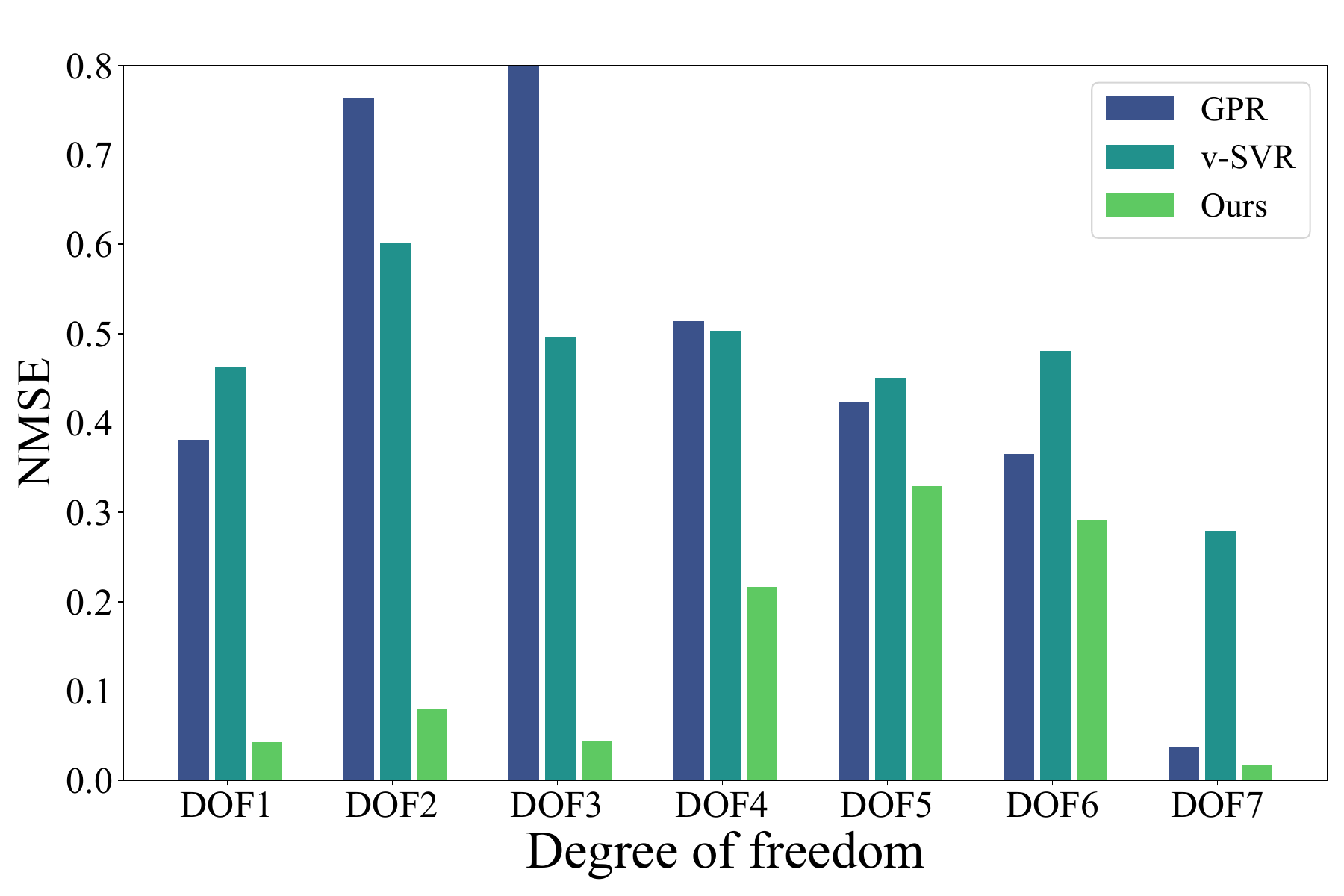}
  \caption{Accuracy on the iCub Robot}
  \label{pic_icubnmse}
\end{figure}

\begin{figure*}[htbp] 
    \centering
    \begin{minipage}[b]{0.48\textwidth}
        \centering
        \includegraphics[width=\textwidth]{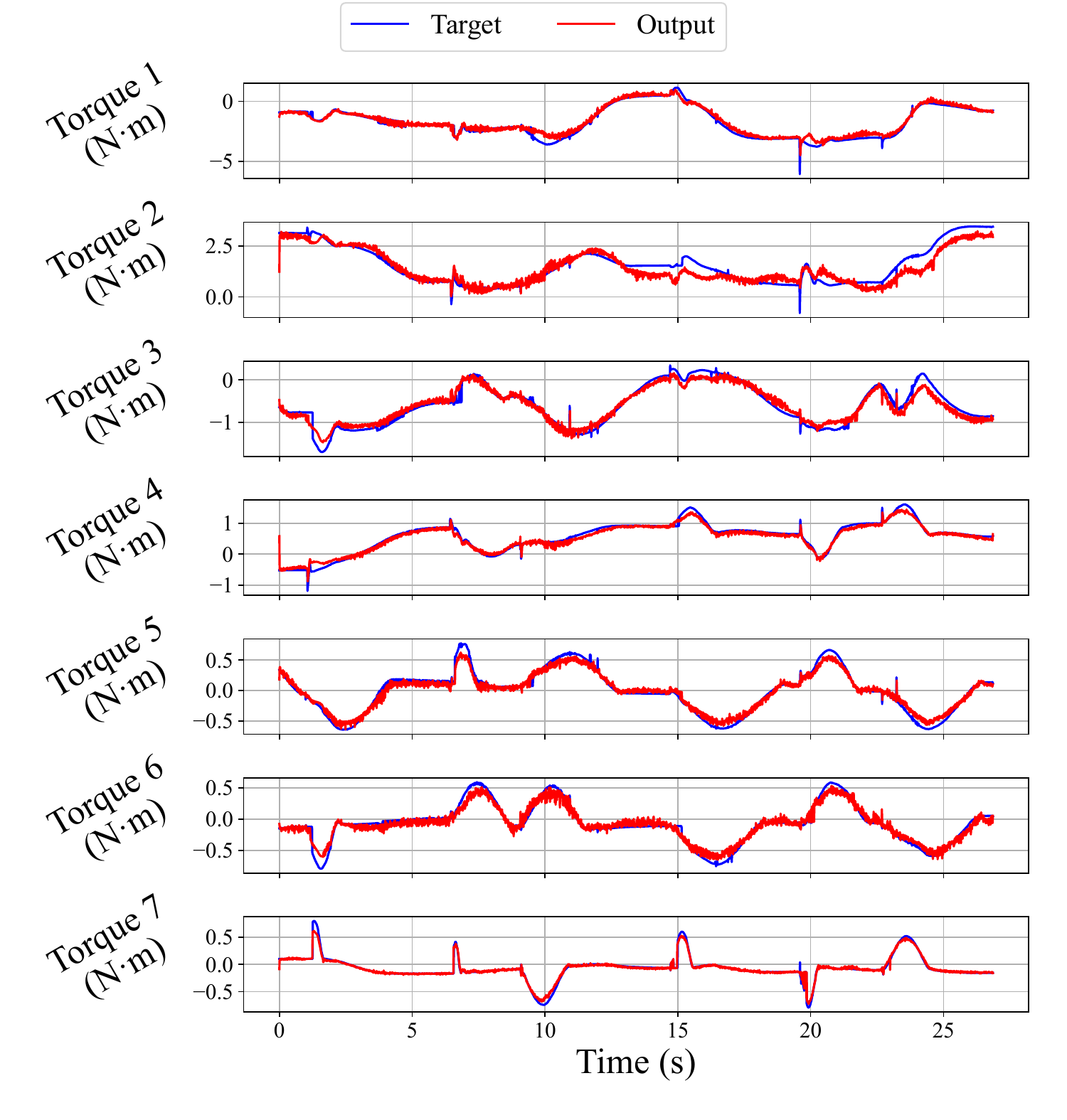}
         \subcaption{Joint Torque}
    \end{minipage}
    \hfill
    \begin{minipage}[b]{0.48\textwidth}
        \centering
        \includegraphics[width=\textwidth]{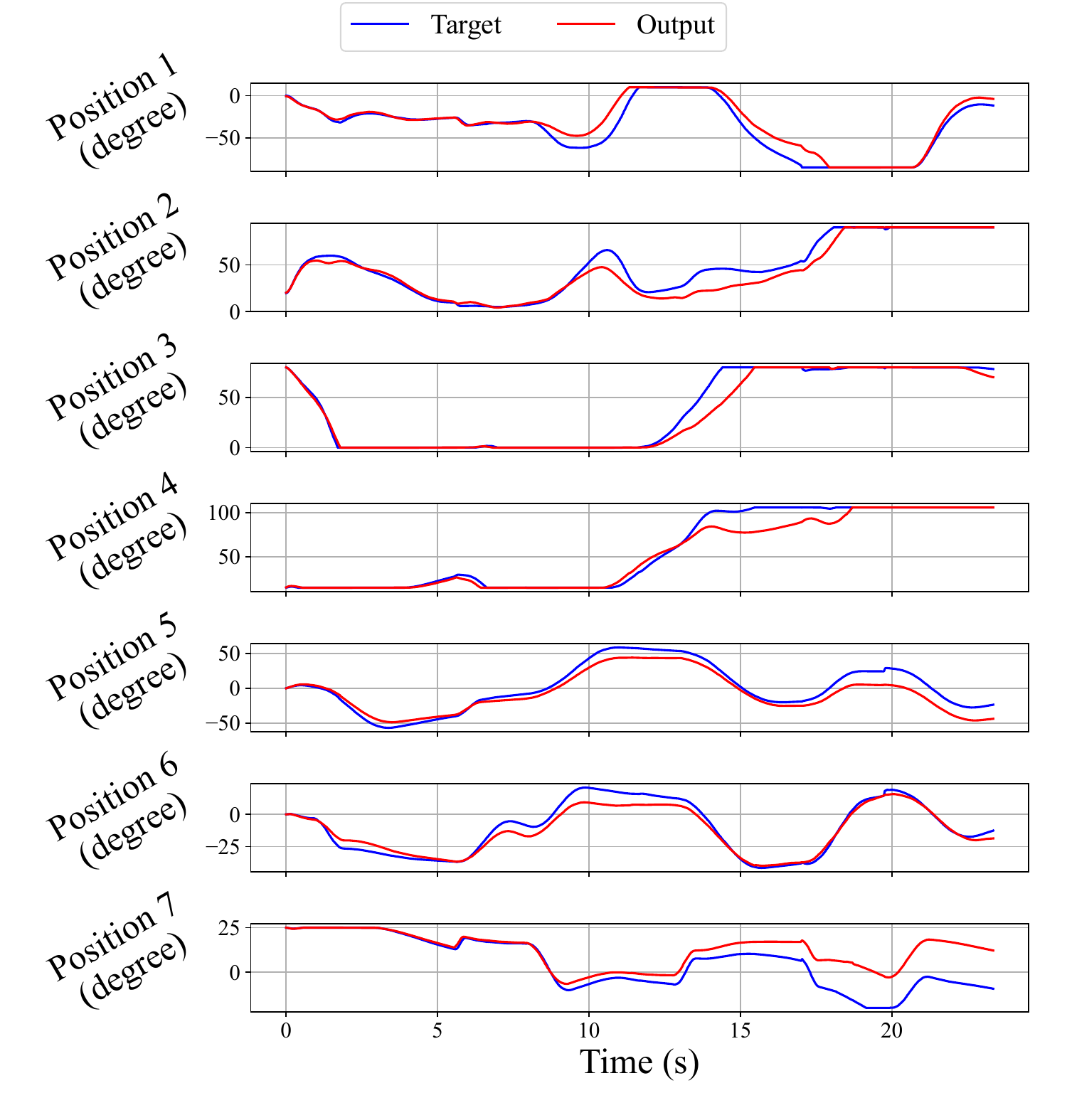}
        \subcaption{Joint Trajectory}
    \end{minipage}
    \caption{Comparison results of iCub robotic arm trajectory tracking. (a) The comparison between the desired target torque required by the iCub robot's predefined ideal trajectory and the output torque predicted by the network. (b) The comparison between the target position and the output position during the actual motion of the iCub robot.}
    \label{combined_images}
\end{figure*}

We also carried out experiments on the iCub robot. In Fig. \ref{pic_icubnmse} the results show that the torque predicted by our method is more accurate. At the same time, we apply the proposed algorithm to the robot. The algorithm predicts the corresponding torque according to the preset trajectory, and the robot arm moves according to the predicted torque. We can see in Fig. \ref{combined_images} that the robot can track the desired trajectory successfully.

\subsection{Impact of Spike Encoding on Accuracy}
We further explored the effects of spike encoding. Table \ref{delta} shows the errors obtained from three encoding schemes in two platforms. Based on these results, it can be deduced that delta encoding can capture more dynamic changes and better encode spatiotemporal information.

\begin{table}[h]
    \centering
    \caption{Impact of Spike Encoding}
    \setlength{\tabcolsep}{6pt} 
    \renewcommand{\arraystretch}{1.5} 
    \label{table_result1}
    \begin{tabular}{c|ccc}
        \toprule   
        \multirow{2}{*}{Dataset} & \multicolumn{3}{c}{MSE} \\ 
        \cline{2-4}
        & Delta & Direct & Rate\\  
        \midrule   
        BaxterRand & \textbf{0.0177} & 0.0204 & 0.0183\\  
        \midrule
        iCub &  \textbf{0.146} & 0.229 & 0.149 \\
        \bottomrule 
    \end{tabular}
    \label{delta}
\end{table}

\subsection{Effect of PSO Search on Accuracy}
\begin{figure}[h]
  \centering
  \includegraphics[width=0.48\textwidth]{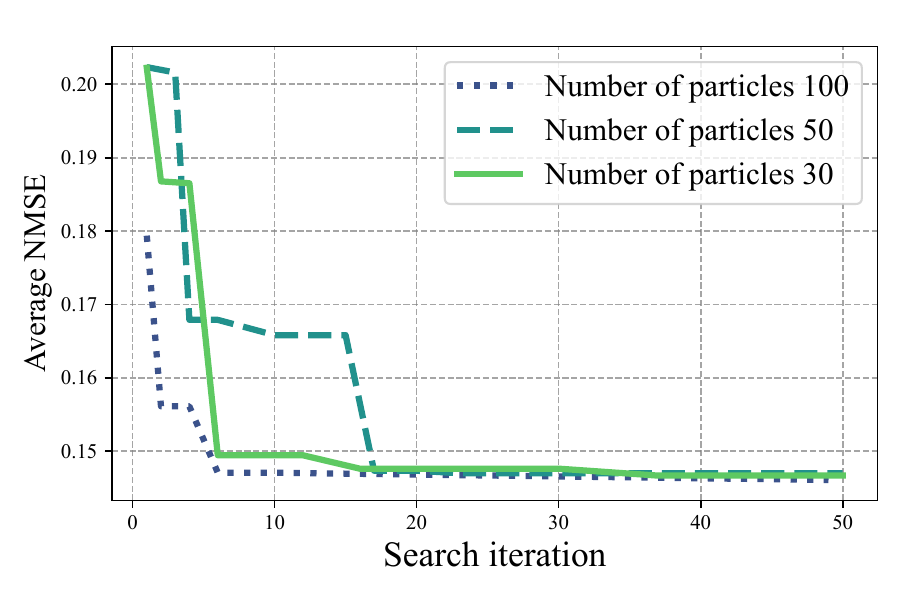}
  \caption{Effect of the PSO search on the error of torques' estimation for the proposed algorithm. The three lines represent different amounts of random particles used in the search.}
  \label{pic_search}
\end{figure}

The structure of the LSM also affects the performance of the network. We used the PSO method to search for the optimal LSM structure. The search results are plotted in  Fig. \ref{pic_search}. As can be seen, better-performing LSM structures can be found as the number of search iterations increases. Additionally, We found that using a larger number of particles during the search process can accelerate the error reduction in the early stages of the search.

\section{CONCLUSIONS}
In this paper, we propose a neuromorphic approach to address the inverse dynamics control problem for a 7-DOF robotic arm. Our method not only reduces extensive manual parameter tuning and  high computational costs of traditional model learning algorithms, but also increases the degrees of freedom an SNN-based algorithm can control. Specifically, we use delta encoding to effectively capture the input changes and reduce the average NMSE by 36\%. We design an LSM-based structure to exploit the inherent spatiotemporal feature in the robot movements. Our control framework reduce the average NMSE of 7-DOF torque prediction by 60\% and is demonstrated in a targel-tracking task. This work advances the development of embodied intelligence with neuromorphic computing technology. In the future, this network can be implemented on neuromorphic hardware thanks to its hardware-friendly SNN structure and deployed on embodied systems to build truly autonomous agents.
\addtolength{\textheight}{-10.5cm}   

\bibliographystyle{splncs04}
\bibliography{ieeeconf/reference}

\end{document}